\documentclass[letterpaper]{article} 
\usepackage[draft]{aaai25}  
\usepackage{times}  
\usepackage{helvet}  
\usepackage{courier}  
\usepackage[hyphens]{url}  
\usepackage{graphicx} 
\usepackage{natbib}  
\usepackage{caption} 
\usepackage{algorithm}
\usepackage{algorithmic}

\usepackage{newfloat}
\usepackage{listings}
\DeclareCaptionStyle{ruled}{labelfont=normalfont,labelsep=colon,strut=off} 
\floatstyle{ruled}
\newfloat{listing}{tb}{lst}{}
\floatname{listing}{Listing}
\PassOptionsToPackage{usenames,dvipsnames,table}{xcolor}
\usepackage{amsmath}  
\usepackage{bm}  
\usepackage{subcaption}  
\usepackage{tikz}  
\usetikzlibrary{matrix, positioning}  
\usepackage{pgfplots}  
\usepackage{amsthm}
\usepackage{adjustbox}  
\usepackage{booktabs}

\newtheorem{theorem}{Theorem}
\newtheorem{lemma}{Lemma}
\newtheorem{example}{Example}

\usepackage{amsmath,amsfonts,bm}

\usepackage{amsthm}
\usepackage{mathtools}

\newcommand{\lsbb}[2]{
    \ifthenelse{\equal{#2}{0}}
    {\it{BL}_{#1, #2}}
    {\it{BL}_{#1, #2}}
}
\newcommand{\lsbbcoeff}[2]{
    \ifthenelse{\equal{#2}{0}}
    {C_{\it{BL}_{#1, #2}}}
    {C_{\it{BL}_{#1, #2}}}
}
\newcommand{\lsbbconst}[2]{
    \ifthenelse{\equal{#2}{0}}
    {c_{\it{BL}_{#1, #2}}}
    {c_{\it{BL}_{#1, #2}}}
}
\newcommand{\rlsbb}[2]{
    \ifthenelse{\equal{#2}{0}}
    {\overline{BL}_{#1, #2}}
    {\overline{BL}_{#1, #2}}
}
\newcommand{\rlsbbcoeff}[2]{
    \ifthenelse{\equal{#2}{0}}
    {\overline{C}_{\it{BL}_{#1, #2}}}
    {\overline{C}_{\it{BL}_{#1, #2}}}
}
\newcommand{\rlsbbconst}[2]{
    \ifthenelse{\equal{#2}{0}}
    {\overline{c}_{\it{BL}_{#1, #2}}}
    {\overline{c}_{\it{BL}_{#1, #2}}}
}

\newcommand{\usbb}[2]{
    \ifthenelse{\equal{#2}{0}}
    {\it{BU}_{#1, #2}}
    {\it{BU}_{#1, #2}}
}
\newcommand{\usbbcoeff}[2]{
    \ifthenelse{\equal{#2}{0}}
    {C_{\it{BU}_{#1, #2}}}
    {C_{\it{BU}_{#1, #2}}}
}
\newcommand{\usbbconst}[2]{
    \ifthenelse{\equal{#2}{0}}
    {c_{\it{BU}_{#1, #2}}}
    {c_{\it{BU}_{#1, #2}}}
}

\def\eqref#1{equation~\ref{#1}}

\def\1{\bm{1}}

\def\va{{\bm{a}}}
\def\vb{{\bm{b}}}

\def\vd{{\bm{d}}}

\def\vr{{\bm{r}}}

\def\vw{{\bm{w}}}
\def\vx{{\bm{x}}}
\def\vy{{\bm{y}}}

\def\mA{{\bm{A}}}
\def\mB{{\bm{B}}}
\def\mC{{\bm{C}}}

\def\mI{{\bm{I}}}

\def\mK{{\bm{K}}}
\def\mL{{\bm{L}}}

\def\mP{{\bm{P}}}

\def\mR{{\bm{R}}}

\def\mU{{\bm{U}}}

\def\mW{{\bm{W}}}

\DeclareMathAlphabet{\mathsfit}{\encodingdefault}{\sfdefault}{m}{sl}
\SetMathAlphabet{\mathsfit}{bold}{\encodingdefault}{\sfdefault}{bx}{n}

\def\sN{{\mathbb{N}}}

\DeclareMathOperator*{\argmax}{arg\,max}

\title{Verification of Neural Networks Against Convolutional
  Perturbations \\
via Parameterised Kernels}
\author {
    Benedikt Brückner\textsuperscript{\rm 1,2},
    Alessio Lomuscio\textsuperscript{\rm 1,2}
}
\affiliations {
    \textsuperscript{\rm 1}Safe Intelligence\\
    \textsuperscript{\rm 2}Imperial College London\\
    benedikt@safeintelligence.ai, alessio@safeintelligence.ai
}

\begin{document}

\maketitle

\begin{abstract}
We develop a method for the efficient verification of neural networks
against convolutional perturbations such as blurring or sharpening. To
define input perturbations, we use well-known camera shake, box blur
and sharpen kernels. We
linearly parameterise these kernels in a way
that allows for a variation of the perturbation strength while
preserving desired kernel properties. To facilitate their use in
neural network verification, we develop an efficient way of convolving
a given input with the parameterised kernels.
The result of this convolution can be used to encode the perturbation
in a verification setting by prepending a linear layer to a given
network. This leads to tight bounds and a high effectiveness
in the resulting verification step. We add further precision by
employing input splitting as a branching strategy. We
demonstrate that we are able to verify robustness on a number
of standard benchmarks where the baseline is unable to provide
any safety certificates. To the best of our knowledge, this is
the first solution for verifying robustness against specific convolutional
perturbations such as camera shake.
\end{abstract}

%

\section{Introduction}
\label{sec:introduction}
As neural networks are increasingly deployed in safety-critical
domains such as autonomous vehicles, aviation, or robotics, concerns about
their reliability are rising. Networks have been shown to be vulnerable to
\textit{Adversarial Attacks}, i.e., perturbations that are often imperceptible,
but change the output of a model for a given
instance \cite{Madry+17}. Such adversarial examples have
been shown to also exist in the physical world and pose a threat to
algorithms deployed in practical applications \cite{Tu+20}.
Neural Network Verification has been put forward as a way to address
these issues by formally establishing that for a given input, a
network is robust with respect to a set of specified perturbations;
this property is often referred to as \textit{local
robustness} \cite{Katz+17,Gehr+18a,Singh+18}.

Verification algorithms are usually divided into complete and
incomplete approaches. Given enough time, complete methods are
guaranteed to provide a definitive answer to the verification
problem. In contrast, incomplete methods may not be able to answer the
verification problem, returning an \textit{undecided} result. Complete
approaches often employ an exact encoding of the network at hand. They
rely on techniques such as Mixed Integer Linear Programming (MILP)
\cite{TjengXiaoTedrake19,Anderson+20,Bunel+20} or Satisfiability Modulo
Theories (SMT) \cite{PulinaTacchella12,Katz+17}.  Incomplete verifiers
employ methods such as Semidefinite Programming
\cite{RaghunathanSteinhardtLiang18,Batten+21,LanBruecknerLomuscio23}, or
bound propagation \cite{Wang+18a,Wang+18b,Singh+19a,Xu+21,Wang+21b}.
They usually overapproximate the true behaviour of the neural network
but can be made complete by combining them with a Branch and Bound
(BaB) strategy.  Stronger verifiers either employ tighter relaxations
such as SDP-based ones or linear constraints that reason over multiple
neurons simultaneously
\cite{Singh+19,Mueller+21,Ferrari+22,Zhang+22}. State-of-the-art (SoA) verifiers
achieve low runtimes through exploiting GPU-enabled
parallelism \cite{Brix+23b, Brix+23a}.

Early approaches established local robustness against norm-based
perturbations, often referred to as
\textit{white noise}~\cite{PulinaTacchella12,Singh+18,Katz+19}. A number of
other perturbations were later proposed. Robustness to photometric perturbations
such as brightness, contrast, hue or saturation changes as well as
more expressive
\textit{bias field} perturbations can be verified by
prepending suitable layers to a neural network
~\cite{KouvarosLomuscio18,Henriksen+21,Mohapatra+20}.
Verifiers were also extended to handle complex geometric
perturbations such as rotations, translations, shearing or scaling,
although the efficient verification against such perturbations
requires further modifications and extensions
\cite{KouvarosLomuscio18,Singh+19a,Balunovic+19a,Batten+24}.
Other approaches focus on the efficient verification of robustness to
occlusions
\cite{Mohapatra+20,Guo+23} or semantically rich perturbations in the latent
space of generative models \cite{Mirman+21,HanspalLomuscio23}.

More relevant to this work are previous investigations of camera shake
effects. \citet{Guo+20} examine the performance of networks in the
presence of motion blur. The results show a significant degradation of
the performance of the models. This phenomenon can be modeled by
applying a convolution operation using suitable kernels to a given
input image~\cite{Sun+15,Mei+19}. Using different kernels, convolution
operations can similarly be used to implement other image
transformations which include box
blur \cite[pp.153--154]{ShapiroStockman01} and sharpen
\cite[pp.50--56]{Arvo91}. Since many semantically interesting and realistic
perturbations can be modelled by using convolutions, being able to
verify robustness to perturbations in a kernel space is highly
valuable. Some attempts at verifying the robustness of models to such
perturbations have been made before. \citet{Paterson+21} encode
contrast, haze and blur perturbations but only perform verification
for haze while resorting to empirical testing for contrast and
blur. One previous approach presents a general method which, if
successful, certifies robustness of a network to all possible
perturbations represented by a kernel of the given
size \cite{MziouSallamiAdjed22}.  However, this generality comes at a
cost. It leads to loose bounds and a high dimensionality of the
perturbation which makes verification difficult, even more so for
large networks. The universality also implies that counterexamples
which are misclassified by the network may be difficult to interpret.

In this work we propose a new method which aims at the tight
verification of networks against convolutional perturbations with a
semantic meaning. Our key contributions are the following:
\begin{itemize}
\item We present an efficient, symbolic encoding of the perturbations
and show how an arbitrary but constant input can efficiently be convolved
with a linearly parameterised kernel using standard convolution
operations.
\item We present parameterised kernels for motion blur
perturbations with various blurring angles as well as box blur and sharpen.
\item Using standard benchmarks from past editions of the Verification of
Neural Networks Competition \cite{Brix+23b,Brix+23a} as well as
self-trained models, we show
experimentally that verification is significantly
easier with our method due to the tighter bounds and the low dimensionality
of the perturbation. Our ablation study demonstrates that the existing
method is unable to verify any properties on the networks we use. Our
method certifies a majority of the properties for small
kernel sizes and perturbation strengths while still being able to certify
robustness in a number of cases for large kernel sizes and strengths.
\end{itemize}

\section{Background}
\label{sec:background}
The notation used for the remainder of the paper is as follows: we use bold
lower case letters $\va$ to denote vectors with $\va[i]$ representing the
$i$-th element of a vector, bold upper case letters $\mA$ to denote
matrices with $\mA[i,j]$ to denote the element in the $i$-th row and $j$-th
column of a matrix and $\left \| \va \right \|_\infty := \sup_i |\vx[i]|$
for the $l_\infty$ norm of a vector. Since we focus on neural networks for
image processing, we assume that the input of a given network will be an
image and therefore refer to single entries in an input matrix as pixels.

\subsection{Feed-forward Neural Networks}
\label{ssec:feed_forward_neural_networks}
A feed-forward neural network
(FFNN) is a function $f(\vx): \mathbb{R}^{n_0} \rightarrow
\mathbb{R}^{n_L}$ which is defined using the concatenation of $L \in
\mathbb{N}$ \textit{layers}. Each layer itself implements a function $f_i$
and it holds that $f(\vx) = f_L(f_{L-1}(\dots(f_1(\vx))))$. Given an input
$\vx_0$ the output of a layer $1 \leq i \leq L$ is calculated in a
recursive manner by applying the layer's operations to the output of the
previous layer, i.e. $\vx_i = f_i(\vx_{i-1})$. The operation encoded by the
$i$-th layer is $f_i: \mathbb{R}^{n_{i-1}} \rightarrow \mathbb{R}^{n_i}$
where $n_i$ is the number of \textit{neurons} in that layer. We assume each
layer operation $f_i$ to consist of two components: firstly, the
application of a linear map $a_i: \mathbb{R}^{n_{i-1}} \rightarrow
\mathbb{R}^{n_i}, \vx_{i-1} \mapsto \mW_i \vx_{i-1} + \vb_i$ for a weight
matrix $\mW_i \in \mathbb{R}^{n_i \times n_{i-1}}$ and a bias vector $\vb_i
\in \mathbb{R}^{n_i}$ which yields the pre-activation vector
$\hat{\vx}_{i}$. And secondly the element-wise application of an
\textit{activation function} $\sigma_i: \mathbb{R}^{n_i} \rightarrow
\mathbb{R}^{n_i}$ yielding the post-activation vector $\vx_i$. In the
verification literature networks are often assumed to use the piece-wise
ReLU function $\text{ReLU}(x) = \max(0, x)$, but verification is equally
possible for other activations like sigmoid or tanh functions
\cite{Shi+24}. The last layer of a given network normally does not include
an activation function, $\sigma_L$ would therefore be the identity map. In
this work we focus on networks performing image classification where the
input $\vx_0$ is an image that needs to be categorised as belonging to one
out of $c$ classes. The final layer outputs $n_L = c$ classification scores
and the predicted class for an image is $j = \argmax_i x_L[i] $.

\subsection{Neural Network Verification}
\label{ssec:neural_network_verification}
Given a trained network $f$, the
verification problem consists of formally showing that the output of the
network is always contained in a linearly definable set $\mathcal{O}
\subset \mathbb{R}^{n_L}$ for all inputs in a linearly definable input set
$\mathcal{I} \subset \mathbb{R}^{n_0}$. Formally we aim to show that
\begin{displaymath}
\forall \vx_0 \in \mathcal{I}: f(\vx_0) \in \mathcal{O}
\end{displaymath}
Inspired by adversarial attack
paradigms, most works study the local robustness of networks to white noise
contrained by the $l_\infty$ norm \cite{Bastani+16,Singh+18}. Given an
input $\bar{\vx} \in \mathbb{R}^{n_0}$ which the network correctly
classifies as belonging to class $j'$, they do so by defining the input and
output sets as
\begin{align*}
\mathcal{I} &= \left \{ \vx \in \mathbb{R}^{n_0} \mid \|\vx - \bar{\vx} \|_\infty \leq \epsilon \right \} \\
\mathcal{O} &= \left \{ \vy \in \mathbb{R}^{n_L} \mid \vy[j'] > \vy[j] \forall j \neq j' \right \}
\end{align*}
where $\epsilon$ is the
perturbation size for which the verification query should be solved.

SoA verifiers often employ bound propagation of some
kind \cite{Liu+19a,Meng+22}.  If the bounds obtained at the final
layer are tight enough, they can be used to answer the verification
problem. The key difficulty in these approaches lies in the
nonlinearity of the network activation functions.  Convex relaxations
of the functions are usually employed, but they induce an
overapproximation error which can become significant for larger
networks \cite{Liu+19a}. Most methods employ a BaB mechanism which
allows for a refinement of the network encoding if the problem cannot
be solved with the initial encoding due to the relaxations being too
coarse
\cite{DePalma+21,Ferrari+22,Shi+24}. One branching
strategy is \textit{input splitting} which partitions the input space
into subspaces and has been found to be particularly effective for
networks with low input dimensions
\cite{Wang+18b,Botoeva+20}.
\textit{Neuron splitting} is also used for
networks with high-dimensional perturbations where input splitting is
less effective \cite{Botoeva+20}. This strategy splits the input space
of a single neuron in the network into subspaces to allow for a more
precise encoding of the activation functions. In the simple case of
piece-wise ReLU activation functions, this can be done by splitting
the function into its two linear pieces \cite{Ferrari+22}.
Verification for high-dimensional perturbations such as norm-based
ones is normally more challenging than verification for
low-dimensional properties like brightness or
contrast \cite{Wang+18a}.  This is due to the \textit{dimensionality}
of the perturbations and the BaB strategies mentioned before.
The typical return values of the verifiers consist of
either \textit{safe} (the network is robust under the given
perturbation), \textit{undecided} (the verifier could neither verify
nor falsify the query, for example due to overly coarse relaxations),
or \textit{unsafe} (a concrete counterexample for which the network
returns an incorrect result was found in the space of allowed
perturbations).

\subsection{Convolution}
\label{ssec:convolution}
\begin{figure*}[!htb]
\centering
\small
\begin{subfigure}{0.245\textwidth}
\centering
\begin{displaymath}
\begin{pmatrix}
0 & 0 & 0 \\[2pt]
0 & 1 & 0 \\[2pt]
0 & 0 & 0 \\
\end{pmatrix}
\end{displaymath}
\resizebox{.8\linewidth}{!}{
\includegraphics[width=\linewidth]{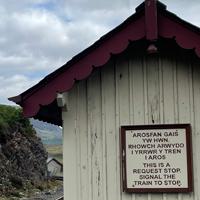}
}
\caption{Identity}
\end{subfigure}
\begin{subfigure}{0.245\textwidth}
\centering
\begin{displaymath}
\begin{pmatrix}
\frac{1}{9} & \frac{1}{9} & \frac{1}{9} \\[2pt]
\frac{1}{9} & \frac{1}{9} & \frac{1}{9} \\[2pt]
\frac{1}{9} & \frac{1}{9} & \frac{1}{9} \\
\end{pmatrix}
\end{displaymath}
\resizebox{.8\linewidth}{!}{
\includegraphics[width=\linewidth]{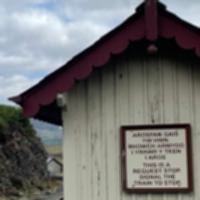}
}
\caption{Box Blur}
\end{subfigure}
\begin{subfigure}{0.245\textwidth}
\centering
\begin{displaymath}
\begin{pmatrix}
0 & -\frac{1}{4} & 0 \\[2pt]
-\frac{1}{4} & 2 & -\frac{1}{4} \\[2pt]
0 & -\frac{1}{4} & 0 \\
\end{pmatrix}
\end{displaymath}
\resizebox{.8\linewidth}{!}{
\includegraphics[width=\linewidth]{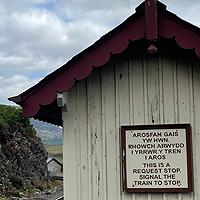}
}
\caption{Sharpen}
\end{subfigure}
\begin{subfigure}{0.245\textwidth}
\centering
\begin{displaymath}
\begin{pmatrix}
0 & 0 & \frac{1}{3} \\[2pt]
0 & \frac{1}{3} & 0 \\[2pt]
\frac{1}{3} & 0 & 0 \\
\end{pmatrix}
\end{displaymath}
\resizebox{.8\linewidth}{!}{
\includegraphics[width=\linewidth]{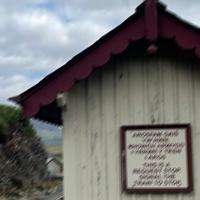}
}
\caption{Motion Blur, $\phi=45^\circ$}
\end{subfigure}
\caption{Visualisation of the basic kernels used in this work}
\label{fig:basic_kernels}
\end{figure*}

\textit{Convolution} is a
mathematical operation which is frequently used for processing inputs in
signal processing. Most relevant for us is the discrete convolution
operation on two-dimensional inputs, in our case images. For a given input
matrix, it computes each element in the output matrix by multiplying the
corresponding input value and its neighbours with different weights and
then summing over the results. Given a two-dimensional input matrix $\mI$
and a \textit{kernel} matrix $\mK$ we define the convolution of $\mI$ with
$\mK$, often written as $\mI * \mK$, as follows
\cite[pp.331--334]{GoodfellowBengioCourville16}:
\begin{displaymath}
\mI * \mK [i,j] = \sum_{k, l} \mI[i+k, j+l] \mK[k, l]
\end{displaymath}
Here $(i, j)$ is a tuple of valid indices for the result, the output shape of
$\mI * \mK$ can be computed based on a number of parameters
\cite{DumoulinVisin18}.
Signal processing literature often assumes that the kernel is flipped
for convolution, but in line with common machine learning frameworks
we omit this and refer to the above operation as convolution. The
elements of $\mK$ are usually normalised to sum to $1$, for example if
$\mI$ is an image, since this preserves the brightness of the
image. The output of a convolution is usually of a smaller size than
the input, if an output of identical size is
required, \textit{padding} can be added to the image before the
convolution.
In early image processing algorithms linear filtering methods such as
convolutions were frequently used for purposes such as edge detection or
denoising with kernels being carefully designed by
experts \cite[pp.100--101]{Szeliski22}. Convolutional neural networks learn
these kernels from data in order to perform a variety of tasks such as
classification or object detection. When using suitable kernels,
convolution can be used to apply effects such as sharpening to an image
[\citealp{Mei+19}, \citealp[p.101]{Szeliski22}]. We focus on the
verification against camera shake or motion blur while also
considering box blur and sharpen to demonstrate the generalisation of our
method to other perturbations. Basic kernels for box blur
\cite[pp.153--154]{ShapiroStockman01}, sharpen \cite[pp.50--56]{Arvo91} and
camera shake \cite{Sun+15,Mei+19} are given in Figure
\ref{fig:basic_kernels} together with the identity kernel. The effect each
of those kernels has on an image is also shown in the figure. Note that for
inputs with multiple colour channels the convolution is performed
independently for each channel.

\subsection{Convolutional Perturbations}
\label{ssec:convolutional_perturbations}
\citet{MziouSallamiAdjed22} propose an algorithm for verifying the robustness
of a network to convolutional perturbations. They define the
neighbourhood of a pixel in the input space for a given kernel size $k$,
input image $\mI \in \mathbb{R}^{d_1
\times d_2}$ and a pixel location tuple $(i, j)$
as the set of all pixels inside a box of size $k\times k$ centered at the
position $(i, j)$. Fields of the box that lie outside the bounds of
the input image are disregarded. A lower bound $l$ and upper bound $u$ for
each pixel is calculated as the minimum and maximum element in that
neighbourhood, respectively. These bounds are tight in the sense that they
are attainable: the lower bound for a pixel can be realised through a
kernel which has a $1$ entry at the location of the minimum in the pixel's
neighbourhood and zero elsewhere. A similar construction is possible for
the maximum. If this operation is repeated for each pixel in the input
image, one obtains two matrices $\mL \in \mathbb{R}^{d_1 \times d_2}$ and
$\mU \in \mathbb{R}^{d_1 \times d_2}$ with lower and upper bounds for each
pixel. A standard verifier can be used to certify robustness for a network
on the given input by treating the perturbation as an $l_\infty$, assuming
that each pixel can vary independently between its lower and upper bound.

If verification is successful, the network is certified to be robust to any
convolutional perturbation for which the values of the kernel lie in the
$[0, 1]$ interval. However, since the robustness specification is very
general, the bounds which are obtained from the method can be extremely
loose. The assumption that variations of pixels are not coupled also means
that the perturbation is high dimensional, leading to long runtimes and
even looser bounds in layers deeper in the network. The combination of
these factors means that even for the smallest kernel size of $3$,
\citet{MziouSallamiAdjed22} obtain a verified accuracy of $30\%$ in the
best case and $0\%$ in the worst case for small classifiers trained on the
MNIST and CIFAR10 datasets. The approach is therefore unlikely to scale to
larger networks.

\section{Method}
\label{sec:method}
Phenomena such as motion blur cannot be modeled using standard techniques such as $l_\infty$
perturbations since the computation of each output pixel's value is based not
only on its original value, but also on the values of its neighbouring
pixels. We parameterise specific kernels to model perturbations using convolution,
allowing for the certification of robustness to specific types of
convolutional perturbations while yielding tighter bounds. A major
advantage of our approach is its simplicity. It can be implemented using
standard operations from a machine learning library to calculate the
parameters of a linear layer prepended to the network to be verified. No
special algorithms as in the case of geometric perturbations are required to perform
efficient verification \cite{Balunovic+19a}.

\subsection{Convolution with Parameterised Kernels}
\label{ssec:convolution_with_parameterised_kernels}
In our work we use linearly parameterised kernels in the convolution
operation.  We refer to a kernel as linearly parameterised if each
entry in the kernel matrix is an affine expression depending on a
number of $m$ variables. When convolving a constant input with such a
kernel, the result is again a linear expression because of the
linearity of the convolution operation.

\begin{theorem} \label{th:separable_convolution}
Assume we are given an input image $\mI$ and a parameterised kernel $\mK$
defined as
\begin{displaymath}
\mK = \sum_{i=1}^m \mA_i \cdot z_i + \mB
\end{displaymath}
where $z_i \in \mathbb{R}$. $\mA_i$ and $\mB$ are a number of coefficient
matrices and a bias matrix, respectively, which have the same shape as
$\mK$.
Then we have:
\begin{equation}
\mI * \mK = \sum_{i=1}^m \left ( \mI * \mA_i \right ) z_i + \mI * \mB \label{eq:convsep}
\end{equation}
Proof: See Appendix \ref{ssec:proof_theorem_separable_convolution}.
\end{theorem}
This theorem allows us to compute the result of a convolution with a
parameterised kernel by separately convolving the input with each
coefficient matrix and the bias matrix (Equation \ref{eq:convsep}).
The $z_i$ variables can be ignored during these computations which
means that the convolution operations need to only be executed on
$(m+1)$ constant matrices. Standard convolution implementations from a
machine learning library can thus be used for this computation.

\subsection{Parameterised Kernels for Modelling Camera Shake}
\label{ssec:parameterised_kernels_for_modelling_camera_shake}
To enable the verification of a network to a range of perturbation
strengths, we model a linear transition from the identity
kernel to a desired perturbation kernel $\mP$ using a
variable $z \in [0, 1]$. Two initial
conditions are given for the kernel: it should be equal to the identity kernel
for $z=0$, and equal to the desired perturbation kernel such as those
in Figure \ref{fig:basic_kernels} for $z=1$. An affine function is
unambiguously defined by these two points (that it intercepts),
and we can compute the slope and intercept for each entry in the
kernel. Example \ref{ex:motion_blur_3x3} shows this derivation for the
$3 \times 3$ motion blur kernel with a blur angle of $\phi = 45^\circ$
introduced in Section \ref{ssec:convolution}.

\begin{example} \label{ex:motion_blur_3x3}
The initial conditions for our parameterisation are:
\begin{displaymath}
\mP_{z=0} = \begin{pmatrix}
0 & 0 & 0 \\[2pt]
0 & 1 & 0 \\[2pt]
0 & 0 & 0 \\
\end{pmatrix},
\mP_{z=1} = \begin{pmatrix}
0 & 0 & \frac{1}{3} \\[2pt]
0 & \frac{1}{3} & 0 \\[2pt]
\frac{1}{3} & 0 & 0 \\
\end{pmatrix}
\end{displaymath}
We assume that each kernel entry is of the form $p(z)
= az + b$ where $z \in \mathbb{R}$ is a variable and $a, b \in \mathbb{R}$
are parameters. Since we have two unknowns $a, b$ and two points that
the function passes through from the initial conditions, we can solve for
$a, b$ to obtain our parameters for each kernel entry. In the camera shake
case for $\phi = 45^\circ$ we have three types of entries: The center
entry, the non-center entries on the antidiagonal running from the top
right to the bottom left of the matrix, and the entries that do not lie on
the antidiagonal.
\paragraph{Parameterisation for the Center Entry.}
For the center entry we have $p(0) = 1$ and $p(1) = \frac{1}{3}$. This
results in the constraints
\begin{align}
p(0) = 1 = a \cdot 0 + b \label{eq:mblur1}\\
p(1) = \frac{1}{3} = a \cdot 1 + b \label{eq:mblur2}
\end{align}
Solving for $a, b$ yields $a=-\frac{2}{3}, b=1$ and therefore
$p(z) =-\frac{2}{3} z + 1$.

\paragraph{Parameterisation for the Antidiagonal Entries.}
For non-center entries on the antidiagonal the initial conditions are
\begin{align}
p(0) = 0 = a \cdot 0 + b \label{eq:mblur3}\\
p(1) = \frac{1}{3} = a \cdot 1 + b \label{eq:mblur4}
\end{align}
Equation \ref{eq:mblur3} implies that we have $b=0$ and then
$a=\frac{1}{3}$ follows from Equation \ref{eq:mblur4}. We obtain $p(z) =
\frac{1}{3} z$

\paragraph{Parameterisation for the Off-Antidiagonal Entries.}
For the entries off the antidiagonal we have $p(0)=p(1)=0$ and therefore
$p(z)=0$.\\

In conclusion, we get $\mP = \mA \cdot z + \mB$ with
\begin{displaymath}
\mA = \begin{pmatrix}
0 & 0 & \frac{1}{3} \\[2pt]
0 & -\frac{2}{3} & 0 \\[2pt]
\frac{1}{3} & 0 & 0 \\
\end{pmatrix},
\mB = \begin{pmatrix}
0 & 0 & 0 \\[2pt]
0 & 1 & 0 \\[2pt]
0 & 0 & 0 \\
\end{pmatrix}
\end{displaymath}
\end{example}

In a similar way to the above, parameterisations can be derived for other
kernel sizes, different motion blur angles and other
perturbations such as blur or sharpen as shown in Figure
\ref{fig:basic_kernels}.
The detailed derivations for those can be found
in Appendix \ref{ssec:generalised_camera_shake_kernels},
Appendix \ref{ssec:derivations_for_parameterised_box_blur_kernels}
and Appendix \ref{ssec:derivations_for_parameterised_sharpen_kernels}.
We do not consider kernel sizes that are even numbers in our experiments
since the identity kernel is not well-defined for
even kernel sizes. An
identity kernel can be approximated by defining the four
entries around the true center of the kernel to have a value of
$\frac{1}{2}$ with all other values being zero. However, this kernel
may still add a noticeable blur to the image.
We still derive
parameterisations for these kernel sizes in case they are of interest to
the reader, but defer those derivations to Appendix
\ref{sec:appendix_even_kernel_sizes}.

\subsection{Integration of Convolutional Perturbations into Neural Network
Verifiers} \label{ssec:integration_into_verifiers}
Given the parameterised kernels from Section
\ref{ssec:parameterised_kernels_for_modelling_camera_shake}, we can use
Theorem \ref{th:separable_convolution} to devise a method for easily
verifying the robustness of neural networks to the perturbations the
kernels encode. Since the parameterised kernels we introduce only depend on
a single variable we omit the indexing of the variable $z$ and the
coefficient matrix $\mA$. We borrow the popular idea of using additional
layers that encode a perturbation which are then prepended to a network for
verification \cite{KouvarosLomuscio18,Mohapatra+20,Guo+23}. Assume a
trained neural network $f$ is given together with a correctly classified
input image in vectorised form $\bar{\vx} \in \mathbb{R}^{o_c \cdot o_h
\cdot o_w}$ where $o_c, o_h, o_w$ are the image's number of channels, height
and width, respectively. We first reshape the vector $\bar{\vx}$ into the
original shape of the image $(o_c, o_h, o_w)$ to obtain an input tensor
$\mI$. This step is necessary for the convolution operation to be
applicable. We then separately convolve $\mI$ with $\mA$ and $\mB$ to
obtain $\mR_\mA := \left ( \mI * \mA \right )$ and $\mR_\mB := \left ( \mI
* \mB \right )$. For inputs with multiple channels each channel is
convolved independently with the same kernels $\mA, \mB$ so the output of
the convolution has the same number of channels as the input.

To encode the perturbation in a network layer, we reshape the resulting
matrices of these convolutions to be vectors $\vr_\mA, \vr_\mB \in
\mathbb{R}^{o_c \cdot o_h \cdot o_w}$ again. We then prepend a matrix
multiplication layer to the network which computes $\tilde{\mA} \cdot z +
\tilde{\mB}$ where $\tilde{\mA} = \vr_\mA, \tilde{\mB} = \vr_\mB$ are
parameters for the layer that are set for each verification query and $z
\in \mathbb{R}$ is the input to the network controlling the strength of the
perturbation. Similarly to the existing approaches which encode properties
to verify in a layer prepended to the network, the image information
is now encoded in the parameters of this new layer.
Despite the fact that all parameterisations used in this work only depend
on one variable, the method generalises to the case where the
number of coefficient matrices in the parameterisation $m$ is greater than
one. In those cases we perform $m+1$ separate convolutions and obtain
$m$ matrices $\mR_{\mA_1}, \dots, \mR_{\mA_m}$ and one matrix $\mR_\mB$
which are reshaped into vectors $\vr_{\mA_1}, \dots, \vr_{\mA_m},
\vr_{\mB}$. Assuming $\vr_{\mA_i} \in \mathbb{R}^{o_c \cdot o_h \cdot o_w}$
are column vectors, they can be concatenated horizontally to form a
parameter matrix $\tilde{\mA} \in \mathbb{R}^{o_c \cdot o_h \cdot o_w, m}$.
The input to the network in this case would be a vector $z \in
\mathbb{R}^{m}$ for parameterising the perturbation. The prepended layer
then computes the matrix-vector product $\tilde{\mA} z$, adds the bias
$\mB$ to it and feeds the resulting vector of size $o_c \cdot
o_h \cdot o_w$ into the first layer of the original network.  The
robustness of the resulting network can be checked by using standard
neural network verifiers. Since the input to the augmented network is
low-dimensional, in our case even just one-dimensional, input
splitting as a branching strategy is particularly effective for
verification \cite{Botoeva+20}.

\section{Evaluation}
\label{sec:evaluation}
\begin{table*}[tb]
\centering
\small
\begin{tabular}{cccccccccccccc}
\toprule
\multicolumn{2}{c}{} & \multicolumn{2}{c}{Box Blur} & \multicolumn{2}{c}{Sharpen} & \multicolumn{2}{c}{Motion Blur $0^\circ$} & \multicolumn{2}{c}{Motion Blur $45^\circ$} & \multicolumn{2}{c}{Motion Blur $90^\circ$} & \multicolumn{2}{c}{Motion Blur $135^\circ$} \\
\cmidrule(r){3-4} \cmidrule(r){5-6} \cmidrule(r){7-8} \cmidrule(r){9-10} \cmidrule(r){11-12} \cmidrule(r){13-14}
$s$ & strength & v/us/to & time & v/us/to & time & v/us/to & time & v/us/to & time & v/us/to & time & v/us/to & time \\
\midrule
3 & 0.2 & 45/0/0 & 87 & 45/0/0 & 80 & 45/0/0 & 78 & 45/0/0 & 77 & 45/0/0 & 78 & 44/1/0 & 78 \\
3 & 0.4 & 44/1/0 & 77 & 45/0/0 & 77 & 45/0/0 & 77 & 45/0/0 & 78 & 45/0/0 & 77 & 44/1/0 & 78 \\
3 & 0.6 & 44/1/0 & 78 & 45/0/0 & 77 & 44/1/0 & 76 & 44/1/0 & 78 & 44/1/0 & 76 & 44/1/0 & 77 \\
3 & 0.8 & 44/1/0 & 80 & 45/0/0 & 79 & 44/1/0 & 77 & 44/1/0 & 79 & 44/1/0 & 76 & 44/1/0 & 81 \\
3 & 1.0 & 44/1/0 & 83 & 45/0/0 & 79 & 44/1/0 & 79 & 44/1/0 & 80 & 44/1/0 & 78 & 42/3/0 & 83 \\
\midrule
5 & 0.2 & 44/1/0 & 75 & 45/0/0 & 76 & 44/1/0 & 77 & 44/1/0 & 76 & 44/1/0 & 76 & 44/1/0 & 76 \\
5 & 0.4 & 44/1/0 & 81 & 45/0/0 & 77 & 44/1/0 & 78 & 44/1/0 & 77 & 44/1/0 & 77 & 44/1/0 & 82 \\
5 & 0.6 & 43/2/0 & 87 & 45/0/0 & 81 & 44/1/0 & 80 & 44/1/0 & 83 & 44/1/0 & 78 & 38/7/0 & 85 \\
5 & 0.8 & 41/4/0 & 91 & 45/0/0 & 84 & 43/2/0 & 83 & 44/1/0 & 88 & 44/1/0 & 82 & 35/10/0 & 87 \\
5 & 1.0 & 36/9/0 & 91 & 45/0/0 & 88 & 42/3/0 & 84 & 41/4/0 & 90 & 44/1/0 & 87 & 31/14/0 & 89 \\
\midrule
7 & 0.2 & 44/1/0 & 77 & 45/0/0 & 76 & 44/1/0 & 77 & 44/1/0 & 78 & 44/1/0 & 77 & 44/1/0 & 76 \\
7 & 0.4 & 42/3/0 & 88 & 45/0/0 & 77 & 44/1/0 & 80 & 44/1/0 & 82 & 44/1/0 & 78 & 38/7/0 & 82 \\
7 & 0.6 & 34/11/0 & 90 & 45/0/0 & 86 & 42/3/0 & 84 & 41/4/0 & 89 & 44/1/0 & 84 & 34/11/0 & 87 \\
7 & 0.8 & 25/20/0 & 87 & 44/1/0 & 88 & 37/8/0 & 87 & 37/8/0 & 91 & 42/3/0 & 88 & 24/21/0 & 85 \\
7 & 1.0 & 20/25/0 & 86 & 43/2/0 & 90 & 33/12/0 & 86 & 30/15/0 & 91 & 39/6/0 & 90 & 22/23/0 & 87 \\
\midrule
9 & 0.2 & 44/1/0 & 78 & 45/0/0 & 77 & 44/1/0 & 77 & 44/1/0 & 77 & 44/1/0 & 76 & 42/3/0 & 78 \\
9 & 0.4 & 36/9/0 & 86 & 45/0/0 & 82 & 42/3/0 & 82 & 42/3/0 & 86 & 44/1/0 & 83 & 35/10/0 & 84 \\
9 & 0.6 & 24/21/0 & 86 & 44/1/0 & 87 & 36/9/0 & 85 & 34/11/0 & 90 & 40/5/0 & 87 & 26/19/0 & 86 \\
9 & 0.8 & 17/28/0 & 83 & 43/2/0 & 90 & 29/16/0 & 87 & 27/18/0 & 89 & 39/6/0 & 91 & 20/25/0 & 86 \\
9 & 1.0 & 9/36/0 & 77 & 43/2/0 & 94 & 26/19/0 & 87 & 16/29/0 & 83 & 36/9/0 & 93 & 13/32/0 & 82 \\
\bottomrule
\end{tabular}
\caption{Experimental evaluation on mnist\_fc. Each query is run with a
timeout of 1800 seconds. $s$ is the filter size and v/us/to denote the
number of verified/unsafe/timeout instances, respectively.}
\label{tab:mnist_fc}
\end{table*}

To evaluate the method described in the previous section, we
used Venus
\cite{KouvarosLomuscio21}, a robustness verification toolkit that
uses both Mixed-Integer Linear Programming and Symbolic Interval
Propagation to solve verification problems. The toolkit was extended to
process modified vnnlib files as used in the most recent Verification of
Neural Networks Competition (VNNCOMP23) which encoded
the perturbations \cite{Brix+23a}.
Venus uses PyTorch \cite{Paszke+19} for efficient vectorised
computations. We enabled its SIP solver and its
adversarial attacks engine, and used the default settings for anything
else. Our method is implemented in PyTorch and reads a vnnlib file,
builds the parameterised kernels as described in Section
\ref{ssec:parameterised_kernels_for_modelling_camera_shake} and
convolves the input with them using PyTorch's \texttt{Conv2d}
operation. An additional layer encoding the perturbation was prepended
to the network as described in Section
\ref{ssec:integration_into_verifiers}, and verification with input
splitting was run on the augmented network using Venus. The experiments
were conducted on a Fedora 35 server equipped with an AMD EPYC
7453 28-Core Processor and 512GB of RAM.

The performance of robustness verification for the proposed
perturbations was evaluated on three benchmarks from previous editions
of the Verification of Neural Networks Competition
\cite{Mueller+22b}. mnist\_fc is a classification benchmark which
consists of three different networks with 2, 4 and 6 layers with 256
ReLU nodes each that is trained on the MNIST dataset. Oval21 contains
three convolutional networks trained on the CIFAR10 dataset. Two of
them consist of two convolutional layers followed by two
fully-connected layers while the third one has two additional
convolutional layers, the number of network activations ranges from
3172 to 6756. Sri\_resnet\_a is a ReLU-based ResNet with one
convolutional layer, three ResBlocks and two fully-connected layers
trained using adversarial training on the CIFAR10 dataset. For our
resnet18 benchmark, we trained a ResNet18 model from the TorchVision
package (version 0.16.0) on the CIFAR10 dataset. The network has
11.7M parameters. For each VNNCOMP benchmark we took the properties
from VNNCOMP and changed the perturbation type
to one of ours before running the experiments. The timeout for each
query was set to 1800 seconds. For resnet18 we selected 50 correctly
classified instances from the CIFAR10 test set for verification.

\begin{table*}[htb]
\small
\centering
\begin{tabular}{cccccccccccccc}
\toprule
\multicolumn{2}{c}{} & \multicolumn{2}{c}{Box Blur} & \multicolumn{2}{c}{Sharpen} & \multicolumn{2}{c}{Motion Blur $0^\circ$} & \multicolumn{2}{c}{Motion Blur $45^\circ$} & \multicolumn{2}{c}{Motion Blur $90^\circ$} & \multicolumn{2}{c}{Motion Blur $135^\circ$} \\
\cmidrule(r){3-4} \cmidrule(r){5-6} \cmidrule(r){7-8} \cmidrule(r){9-10} \cmidrule(r){11-12} \cmidrule(r){13-14}
$s$ & strength & v/us/to & time & v/us/to & time & v/us/to & time & v/us/to & time & v/us/to & time & v/us/to & time \\
\midrule
3 & 0.2 & 46/4/0 & 4781 & 46/4/0 & 4235 & 47/3/0 & 3924 & 46/4/0 & 4985 & 46/4/0 & 3739 & 46/4/0 & 4950 \\
3 & 0.4 & 45/5/0 & 7156 & 46/4/0 & 6241 & 47/3/0 & 6092 & 45/5/0 & 7286 & 46/4/0 & 5729 & 45/5/0 & 7252 \\
3 & 0.6 & 43/7/0 & 8519 & 46/4/0 & 7677 & 45/5/0 & 7284 & 44/6/0 & 8709 & 45/5/0 & 6895 & 40/10/0 & 7943 \\
3 & 0.8 & 37/13/0 & 8864 & 46/4/0 & 8621 & 44/6/0 & 8358 & 41/9/0 & 9904 & 43/7/0 & 7732 & 36/14/0 & 8398 \\
3 & 1.0 & 32/18/0 & 9211 & 46/4/0 & 9325 & 42/8/0 & 8957 & 40/10/0 & 11442 & 42/8/0 & 8374 & 35/15/0 & 10120 \\
\midrule
5 & 0.2 & 46/4/0 & 6415 & 46/4/0 & 5454 & 47/3/0 & 5733 & 46/4/0 & 6291 & 46/4/0 & 5297 & 45/5/0 & 6190 \\
5 & 0.4 & 42/8/0 & 8594 & 46/4/0 & 7614 & 42/8/0 & 7558 & 45/5/0 & 8976 & 45/5/0 & 7727 & 43/7/0 & 8582 \\
5 & 0.6 & 35/15/0 & 9963 & 46/4/0 & 8942 & 40/10/0 & 8973 & 38/12/0 & 10372 & 43/7/0 & 8958 & 37/13/0 & 9935 \\
5 & 0.8 & 24/26/0 & 9641 & 46/4/0 & 10255 & 33/17/0 & 9452 & 29/21/0 & 10706 & 36/14/0 & 9271 & 26/24/0 & 8919 \\
5 & 1.0 & 22/28/0 & 11857 & 43/7/0 & 10658 & 30/20/0 & 10248 & 26/24/0 & 12164 & 34/16/0 & 10852 & 23/27/0 & 10269 \\
\midrule
7 & 0.2 & 46/4/0 & 7295 & 46/4/0 & 6155 & 44/6/0 & 6179 & 46/4/0 & 7022 & 46/4/0 & 6026 & 45/5/0 & 6846 \\
7 & 0.4 & 42/8/0 & 9919 & 45/5/0 & 8310 & 41/9/0 & 8312 & 43/7/0 & 9798 & 45/5/0 & 8526 & 43/7/0 & 9722 \\
7 & 0.6 & 38/12/0 & 13899 & 45/5/0 & 9647 & 35/15/0 & 9859 & 36/14/0 & 11507 & 40/10/0 & 10043 & 37/13/0 & 11704 \\
7 & 0.8 & 22/28/0 & 11823 & 43/7/0 & 10715 & 28/22/0 & 10372 & 27/23/0 & 11890 & 32/18/0 & 10698 & 28/22/0 & 11971 \\
7 & 1.0 & 14/36/0 & 10464 & 40/10/0 & 11178 & 24/26/0 & 11100 & 21/29/0 & 11747 & 21/29/0 & 8213 & 19/31/0 & 10387 \\
\midrule
9 & 0.2 & 46/4/0 & 7914 & 46/4/0 & 6454 & 45/5/0 & 6822 & 46/4/0 & 7596 & 46/4/0 & 6371 & 45/5/0 & 7507 \\
9 & 0.4 & 43/7/0 & 11115 & 44/6/0 & 8482 & 41/9/0 & 9282 & 43/7/0 & 10660 & 45/5/0 & 9120 & 42/8/0 & 10301 \\
9 & 0.6 & 37/13/0 & 14980 & 43/7/0 & 10028 & 34/16/0 & 11029 & 35/15/0 & 12874 & 40/10/0 & 11312 & 35/15/0 & 12814\\
9 & 0.8 & 22/28/0 & 14201 & 41/9/0 & 11066 & 23/27/0 & 10160 & 27/23/0 & 14035 & 25/25/0 & 9322 & 27/23/0 & 13803 \\
9 & 1.0 & 10/40/0 & 7911 & 40/10/0 & 12457 & 19/31/0 & 10685 & 17/33/0 & 12090 & 17/33/0 & 8698 & 21/29/0 & 13471 \\
\bottomrule
\end{tabular}
\caption{Experimental evaluation on resnet18. Each query was run with
a timeout of 1800 seconds. $s$ is the filter size and v/us/to denote the
number of verified/unsafe/timeout instances, respectively.}
\label{tab:resnet18}
\end{table*}
For each of the benchmarks, we tested perturbations with kernel sizes of $3,
5, 7$ and $9$. We varied the upper bound of the perturbation strength which
we simply denote as \textit{strength} in the following, the lower bound for
the strength is zero. For example, a perturbation strength of $0.4$ means
that for the parameterised kernel, the variable $z$ is allowed to vary
within the interval $[0, 0.4]$. The results for mnist\_fc are presented in
Table \ref{tab:mnist_fc}, those for resnet18 in Table
\ref{tab:resnet18}.
The ones for oval21 can be found in Table \ref{tab:oval21} and those for
sri\_resnet\_a in Table \ref{tab:sri_resnet_a} which we move to Appendix
\ref{sec:experimental_results_on_further_benchmarks} due to space
constraints.

We found that our method scales well to very large networks such as ResNet18
and found that the kernel size and perturbation strength had a much larger
impact on the verifiability of a model than its size. Verification on all
benchmarks was fast due to the low dimensionality of our perturbations.
Verification for small perturbation strengths was successful for nearly all
instances, irrespective of the kernel size $s$. For small kernel sizes such
as $s=3$ we further observed that verification was successful even for very
large strengths. For large kernel sizes and large perturbation strengths,
unsafe cases were more likely to be found which is not surprising given that
the degree of corruption for e.g. box blur with a kernel size of $9$ and a
perturbation strength of $1$ is substantial. The differences in robustness
to different types of perturbations were also noteworthy. While robustness
deteriorated severely for box blur and camera shake when larger kernel
sizes or perturbation strengths were considered, networks retained a high
verified robustness against sharpen perturbations. This intuitively makes
sense. While blurring often induces an information loss which can make it
hard to restore the original information of the image, sharpening
emphasises the image texture and can strengthen edges in the image.
This robustness to large sharpen perturbation strengths could be observed for
both MNIST and CIFAR10.

For mnist\_fc we also observed that the robustness of the network to camera
shake perturbations was highly dependent on the perturbation angle, especially
for a kernel size of $9$. While the networks
were vulnerable to camera shake along the $45^\circ$ and $135^\circ$
axis, they were much more robust to blurring along the $0^\circ$ axis.
Motion blurring along the $90^\circ$ axis affected the networks to the least
degree with verified accuracies still being extremely high for strong
perturbations, even for a kernel size of $9$ and a perturbation strength of
$1.0$. The differences in robustness to different camera shake angles
were also observable for resnet18, oval21 and sri\_resnet\_a, even though
they were less prominent.

\begin{table}[htb]
\centering
\setlength{\tabcolsep}{1mm}
\small
\begin{tabular}{ccccccccc}
\toprule
& \multicolumn{2}{c}{mnist\_fc} & \multicolumn{2}{c}{resnet18} & \multicolumn{2}{c}{oval21} & \multicolumn{2}{c}{sri\_resnet\_a}\\
\cmidrule(r){2-3} \cmidrule(r){4-5} \cmidrule(r){6-7} \cmidrule(r){8-9}
$s$ & v/us/to & time & v/us/to & time & v/us/to & time & v/us/to & time \\
\midrule
3 & 0/45/0 & 70 & 0/50/0 & 133 & 0/30/0 & 258 & 0/71/1 & 7311 \\
5 & 0/45/0 & 70 & 0/50/0 & 133 & 0/30/0 & 49 & 0/72/0 & 113 \\
7 & 0/44/1 & 7268 & 0/50/0 & 133 & 0/30/0 & 2298 & 0/71/1 & 7311 \\
9 & 0/44/1 & 7271 & 0/50/0 & 129 & 0/30/0 & 48 & 0/72/0 & 112 \\
\bottomrule
\end{tabular}
\caption{Ablation Study}
\label{tab:ablation_study}
\end{table}
For baseline comparisons we reimplemented the method presented in
\cite{MziouSallamiAdjed22} and the resulting verification queries were
once again solved by Venus. Since the resulting perturbations are
high-dimensional, we used activation splitting instead of input
splitting as a better performing branching strategy. If the baseline
method verifies robustness for a given model, input and kernel size,
the model is robust to any perturbation that can be encoded with a
kernel of the given size for that input. However, Table
\ref{tab:ablation_study} shows that the baseline perturbations lead to
loose bounds due to the high dimensionality of the perturbations
combined with the already loose bounds for each pixel's value for
large neighbourhoods. Even for a small kernel size of $3$, we found
that no properties could be verified for the networks we consider. In
these cases, bounds became so loose that it is easy for the verifier
to find concrete counterexamples in the majority of cases. The
generality of the perturbation specifications in the baseline is one
of its strengths, but at the same time limits its practical use since
we showed that it does not scale to larger models or input sizes. Our
method is less general and we need to do a separate verification run
for each kernel used for modelling a convolutional
perturbation. However, our approach scales to much larger networks and
therefore enables robustness certification in scenarios where the
baseline fails.

In summary, our findings indicate that the method here presented
enables the verification of networks against a range of perturbations
that can be modeled through parameterised kernels. The method provides
tight bounds, permitting the verification of much larger networks, and
allows for an efficient verification due the low
dimensionality of the perturbations.

\section{Conclusions}
\label{sec:conclusion}
Verification against camera shake and related convolutional
perturbations is important since these phenomena are likely to appear
in the real world. So far, verification against such perturbations
was only possible by using an algorithm that yielded counterexamples
which are difficult to interpret and produced loose bounds, preventing
the robustness verification of larger networks or networks with large
input sizes. The here proposed approach based on parameterised kernels
is easy to implement and allows for the verification of network
robustness to a number of semantically interesting perturbations. We
demonstrated the effectiveness of the method on standard benchmarks
from VNNComp and proved that it is able to verify properties for
networks that cannot be solved by the existing baseline. As we showed,
the method can be used to identify weaknesses of models to specific
types of convolutional perturbations that might otherwise remain
hidden such as motion blurring along axes of a specific angle. Since
it is easy to design parameterised kernels for additional perturbation
types besides those presented, we expect that more convolutional
perturbations of practical interest will be developed in the future,
contributing to more thorough robustness checks for deployed AI
systems.

\section*{Acknowledgements}
\label{sec:acknowledgements}
Benedikt Brückner acknowledges support from the UKRI Centre for
Doctoral Training in Safe and Trusted Artificial Intelligence
(EP/S023356/1]. Alessio Lomuscio acknowledges support from the Royal
  Academy of Engineering via a Chair of Emerging Technologies.

\bibliography{bib}
\clearpage
\appendix

\section{Experimental Results on further benchmarks} \label{sec:experimental_results_on_further_benchmarks}
Due to space constraints we include the detailed evaluation of our method
on the oval21 benchmark in Table \ref{tab:oval21} and that on the
sri\_resnet\_a benchmark in Table \ref{tab:sri_resnet_a} here in the
appendix. The general trends observed on the mnist\_fc and resnet18
benchmarks are also observable on these two benchmarks. We include these
additional results in the appendix to demonstrate that our results are
reproducible in a range of scenarios.

\begin{table*}[htb]
\centering
\small
\begin{tabular}{cccccccccccccc}
\toprule
\multicolumn{2}{c}{} & \multicolumn{2}{c}{Box Blur} & \multicolumn{2}{c}{Sharpen} & \multicolumn{2}{c}{Motion Blur $0^\circ$} & \multicolumn{2}{c}{Motion Blur $45^\circ$} & \multicolumn{2}{c}{Motion Blur $90^\circ$} & \multicolumn{2}{c}{Motion Blur $135^\circ$} \\
\cmidrule(r){3-4} \cmidrule(r){5-6} \cmidrule(r){7-8} \cmidrule(r){9-10} \cmidrule(r){11-12} \cmidrule(r){13-14}
$s$ & strength & v/us/to & time & v/us/to & time & v/us/to & time & v/us/to & time & v/us/to & time & v/us/to & time \\
\midrule
3 & 0.2 & 30/0/0 & 53 & 30/0/0 & 53 & 30/0/0 & 53 & 30/0/0 & 52 & 29/1/0 & 52 & 30/0/0 & 53 \\
3 & 0.4 & 30/0/0 & 54 & 30/0/0 & 53 & 29/1/0 & 53 & 30/0/0 & 54 & 29/1/0 & 52 & 30/0/0 & 53 \\
3 & 0.6 & 27/3/0 & 55 & 29/1/0 & 53 & 26/4/0 & 52 & 26/4/0 & 54 & 29/1/0 & 53 & 29/1/0 & 56 \\
3 & 0.8 & 22/8/0 & 53 & 29/1/0 & 54 & 25/5/0 & 53 & 23/7/0 & 53 & 28/2/0 & 53 & 22/8/0 & 53 \\
3 & 1.0 & 22/8/0 & 54 & 29/1/0 & 55 & 25/5/0 & 54 & 23/7/0 & 54 & 27/3/0 & 54 & 22/8/0 & 54 \\
\midrule
5 & 0.2 & 30/0/0 & 54 & 30/0/0 & 53 & 29/1/0 & 54 & 30/0/0 & 53 & 29/1/0 & 52 & 30/0/0 & 53 \\
5 & 0.4 & 22/8/0 & 52 & 29/1/0 & 53 & 25/5/0 & 54 & 24/6/0 & 53 & 28/2/0 & 53 & 24/6/0 & 53 \\
5 & 0.6 & 21/9/0 & 54 & 29/1/0 & 55 & 23/7/0 & 53 & 21/9/0 & 54 & 26/4/0 & 53 & 22/8/0 & 53 \\
5 & 0.8 & 19/11/0 & 55 & 27/3/0 & 54 & 22/8/0 & 55 & 18/12/0 & 54 & 26/4/0 & 54 & 20/10/0 & 53 \\
5 & 1.0 & 16/14/0 & 56 & 26/4/0 & 55 & 19/11/0 & 55 & 18/12/0 & 58 & 23/7/0 & 54 & 20/10/0 & 56 \\
\midrule
7 & 0.2 & 27/3/0 & 52 & 30/0/0 & 54 & 27/3/0 & 54 & 27/3/0 & 53 & 29/1/0 & 53 & 29/1/0 & 54 \\
7 & 0.4 & 22/8/0 & 53 & 29/1/0 & 54 & 24/6/0 & 54 & 23/7/0 & 54 & 26/4/0 & 54 & 23/7/0 & 53 \\
7 & 0.6 & 19/11/0 & 54 & 28/2/0 & 54 & 20/10/0 & 53 & 19/11/0 & 55 & 25/5/0 & 54 & 20/10/0 & 53 \\
7 & 0.8 & 15/15/0 & 56 & 26/4/0 & 54 & 16/14/0 & 53 & 15/15/0 & 56 & 21/9/0 & 53 & 17/13/0 & 55 \\
7 & 1.0 & 14/16/0 & 56 & 26/4/0 & 55 & 16/14/0 & 56 & 14/16/0 & 58 & 21/9/0 & 55 & 17/13/0 & 57 \\
\midrule
9 & 0.2 & 27/3/0 & 54 & 30/0/0 & 54 & 26/4/0 & 53 & 27/3/0 & 53 & 28/2/0 & 53 & 28/2/0 & 54 \\
9 & 0.4 & 21/9/0 & 53 & 28/2/0 & 53 & 22/8/0 & 53 & 22/8/0 & 53 & 26/4/0 & 53 & 22/8/0 & 53 \\
9 & 0.6 & 17/13/0 & 55 & 28/2/0 & 54 & 17/13/0 & 53 & 16/14/0 & 54 & 22/8/0 & 55 & 18/12/0 & 53 \\
9 & 0.8 & 13/17/0 & 56 & 26/4/0 & 55 & 15/15/0 & 54 & 13/17/0 & 56 & 19/11/0 & 54 & 16/14/0 & 56 \\
9 & 1.0 & 10/20/0 & 56 & 26/4/0 & 56 & 13/17/0 & 55 & 11/19/0 & 56 & 18/12/0 & 54 & 14/16/0 & 57 \\
\bottomrule
\end{tabular}
\caption{Experimental evaluation on oval21. Each query is run with a
timeout of 1800 seconds. $s$ is the filter size and v/us/to denote the
number of verified/unsafe/timeout instances, respectively.}
\label{tab:oval21}
\end{table*}

\begin{table*}[htb]
\small
\centering
\begin{tabular}{cccccccccccccc}
\toprule
\multicolumn{2}{c}{} & \multicolumn{2}{c}{Box Blur} & \multicolumn{2}{c}{Sharpen} & \multicolumn{2}{c}{Motion Blur $0^\circ$} & \multicolumn{2}{c}{Motion Blur $45^\circ$} & \multicolumn{2}{c}{Motion Blur $90^\circ$} & \multicolumn{2}{c}{Motion Blur $135^\circ$} \\
\cmidrule(r){3-4} \cmidrule(r){5-6} \cmidrule(r){7-8} \cmidrule(r){9-10} \cmidrule(r){11-12} \cmidrule(r){13-14}
$s$ & strength & v/us/to & time & v/us/to & time & v/us/to & time & v/us/to & time & v/us/to & time & v/us/to & time \\
\midrule
3 & 0.2 & 39/3/0 & 182 & 69/3/0 & 178 & 70/2/0 & 175 & 70/2/0 & 182 & 69/3/0 & 173 & 69/3/0 & 184 \\
3 & 0.4 & 63/9/0 & 210 & 68/4/0 & 193 & 69/3/0 & 192 & 64/8/0 & 213 & 66/6/0 & 191 & 63/9/0 & 215 \\
3 & 0.6 & 59/13/0 & 235 & 68/4/0 & 230 & 67/5/0 & 218 & 62/10/0 & 247 & 60/12/0 & 206 & 60/12/0 & 241 \\
3 & 0.8 & 57/15/0 & 262 & 68/4/0 & 251 & 64/8/0 & 237 & 61/11/0 & 275 & 60/12/0 & 228 & 54/18/0 & 254 \\
3 & 1.0 & 54/18/0 & 273 & 66/6/0 & 268 & 63/9/0 & 254 & 60/12/0 & 298 & 59/13/0 & 242 & 52/20/0 & 265 \\
\midrule
5 & 0.2 & 67/5/0 & 196 & 69/3/0 & 187 & 70/2/0 & 189 & 69/3/0 & 198 & 68/4/0 & 184 & 66/6/0 & 193 \\
5 & 0.4 & 59/13/0 & 245 & 68/4/0 & 232 & 64/8/0 & 229 & 62/10/0 & 246 & 60/12/0 & 219 & 61/11/0 & 248 \\
5 & 0.6 & 54/18/0 & 275 & 66/6/0 & 260 & 62/10/0 & 264 & 58/14/0 & 287 & 58/14/0 & 250 & 51/21/0 & 262 \\
5 & 0.8 & 46/26/0 & 280 & 64/8/0 & 288 & 59/13/0 & 288 & 51/21/0 & 294 & 52/20/0 & 263 & 42/30/0 & 254 \\
5 & 1.0 & 35/37/0 & 261 & 62/10/0 & 300 & 56/16/0 & 296 & 48/24/0 & 310 & 50/22/0 & 273 & 41/31/0 & 277 \\
\midrule
7 & 0.2 & 69/3/0 & 214 & 69/3/0 & 193 & 69/3/0 & 197 & 69/3/0 & 213 & 69/3/0 & 194 & 65/7/0 & 206 \\
7 & 0.4 & 61/11/0 & 275 & 66/6/0 & 243 & 63/9/0 & 249 & 64/8/0 & 277 & 61/11/0 & 237 & 60/12/0 & 264 \\
7 & 0.6 & 52/20/0 & 289 & 65/7/0 & 281 & 59/13/0 & 285 & 55/17/0 & 295 & 53/19/0 & 260 & 49/23/0 & 274 \\
7 & 0.8 & 31/41/0 & 238 & 62/10/0 & 301 & 53/19/0 & 299 & 44/28/0 & 297 & 45/27/0 & 267 & 38/34/0 & 259 \\
7 & 1.0 & 22/50/0 & 213 & 58/14/0 & 305 & 46/26/0 & 303 & 29/43/0 & 249 & 39/33/0 & 270 & 29/43/0 & 251 \\
\midrule
9 & 0.2 & 69/3/0 & 229 & 69/3/0 & 201 & 68/4/0 & 204 & 69/3/0 & 229 & 69/3/0 & 199 & 65/7/0 & 221 \\
9 & 0.4 & 63/9/0 & 296 & 66/6/0 & 254 & 63/9/0 & 269 & 63/9/0 & 288 & 60/12/0 & 248 & 62/10/0 & 284 \\
9 & 0.6 & 48/24/0 & 294 & 65/7/0 & 300 & 54/18/0 & 285 & 52/20/0 & 301 & 52/20/0 & 269 & 49/23/0 & 289 \\
9 & 0.8 & 23/49/0 & 213 & 59/13/0 & 303 & 42/30/0 & 279 & 29/43/0 & 239 & 40/32/0 & 262 & 32/40/0 & 250 \\
9 & 1.0 & 18/54/0 & 202 & 58/14/0 & 327 & 34/38/0 & 275 & 21/51/0 & 225 & 37/35/0 & 268 & 21/51/0 & 231 \\
\bottomrule
\end{tabular}
\caption{Experimental evaluation on sri\_resnet\_a. Each query is run with
a timeout of 1800 seconds. $s$ is the filter size and v/us/to denote the
number of verified/unsafe/timeout instances, respectively.}
\label{tab:sri_resnet_a}
\end{table*}

\section{Proofs} \label{sec:appendix_proofs}
\subsection{Proof of Theorem \ref{th:separable_convolution}}
\label{ssec:proof_theorem_separable_convolution}
Assume we are a given a kernel which is linearly parameterised as
\begin{displaymath}
\mK = \sum_{i=1}^m \mA_i \cdot z_i + \mB
\end{displaymath}
where $z_i \in \mathbb{R}$. $\mA_i$ and $\mB$ are a number of coefficient
matrices and a bias matrix, respectively, which have the same shape as
$\mK$.
If we we want to compute the result of convolving $\mI$ with $\mK$, we can
exploit the linearity of the convolution operation
\cite[pp.95--96]{Szeliski22} to obtain that
\begin{displaymath}
\mI * \mK = \sum_{i=1}^m \left ( \mI * \mA_i \right ) z_i + \mI * \mB
\end{displaymath}
\begin{proof}
For a kernel $\mK$ that is parameterised as specified above and an input $\mI$
it holds that
\begin{align}
\mI * \mK &= \mI * \left ( \sum_{i=1}^m \left ( \mA_i \cdot z_i \right ) +
\mB \right ) \nonumber\\
&= \mI * \sum_{i=1}^m \left ( \mA_i \cdot z_i \right ) + \mI * \mB
\nonumber\\
&= \sum_{i=1}^m \mI * \left ( \mA_i z_i \right ) + \mI * \mB
\label{eq:separable1}\\
&= \sum_{i=1}^m \left ( \mI * \mA_i \right ) z_i + \mI * \mB
\label{eq:separable2}
\end{align}
\end{proof}
Equation \ref{eq:separable1} is already useful since it allows us to
compute the result of the convolution operation through a number of
separate convolutions which are easier to compute. Equation
\ref{eq:separable2} is the more powerful equation in our case though since
it means the result of a convolution with a parameterised kernel can be
computed by simply convolving the input under consideration with each
coefficient matrix and the bias matrix.

\subsection{Proof of Normalisation for Parameterised Kernels}
\label{ssec:proof_theorem_sum_one}
As outlined before, it is common to normalise kernels used in a convolution
operation so that the sum of all kernel entries is one. We choose our
initial kernels such as the identity kernel for $z=0$ and the target kernel
for $z=1$ to have a sum of one. However, from this it is not immediately
clear whether the kernel entries for any $z \in [0, 1]$ will also sum to one.
To prove that this is the case, we first introduce a lemma which is used
for the proof of the main result
\begin{lemma} \label{le:sum_of_affine_functions}
Assume we are given $h \in \mathbb{N}, h > 1$ affine, vector-valued functions
$g_i$ defined as
\begin{displaymath}
g_i(\vw) = \mC_i \vw + \vd_i
\end{displaymath}
where $\mC_i \in \mathbb{R}^{n_a \times n_b}$ and
$\vw \in \mathbb{R}^{n_b}, \vd \in \mathbb{R}^{n_a}$. 
Let $\mathcal{S}(\vw)$ be the sum of those $h$ functions, i.e.
\begin{displaymath}
\mathcal{S}(\vw) = \sum_{i=1}^h g_i(\vw)
\end{displaymath}
then $\mathcal{S}(\vw)$ is an affine function.
\end{lemma}
\begin{proof}
Let let $h \in \sN, h > 1$ be the number of affine, vector-valued functions
$g_i$ that we consider with $\mathcal{S}$ being their sum as defined in
Lemma \ref{le:sum_of_affine_functions}. We provide a constructive proof
showing that $\mathcal{S}$ is an affine function:
\begin{align*}
\mathcal{S}(\vw) &= \sum_{i=1}^h g_i(\vw) \\
&= \sum_{i=1}^h \left ( \mC_i \vw + \vd_i \right ) \\
&= \sum_{i=1}^h \left ( \mC_i \vw \right ) + \sum_{i=1}^h \vd_i \\
&= \underbrace{\left ( \sum_{i=1}^h \mC_i  \right )}_{\tilde{\mC}} \vw + \underbrace{\sum_{i=1}^h \vd_i}_{\tilde{\vd}} \\
&= \tilde{\mC} \vw + \tilde{\vd}
\end{align*}
where $\tilde{\mC} \in \mathbb{R}^{n_a \times n_b}$ and
$\tilde{\vd} \in \mathbb{R}^{n_a}$ which is again an affine function.
\end{proof}
Theorem \ref{th:sum_one} then shows that if the kernel is normalised for both $z=0$
and $z=1$, then the parameterised kernel $\mP$ derived from
those conditions is also normalised for any $z \in [0, 1]$.
\begin{theorem}\label{th:sum_one}
For any linearly parameterised kernel $\mP(z) \in \mathbb{R}^{k \times k}$
it holds that if
\begin{displaymath}
\sum_{i=1}^k \sum_{j=1}^k \left ( \mP(0)[i, j] \right ) = 1
\end{displaymath}
and
\begin{displaymath}
\sum_{i=1}^k \sum_{j=1}^k \left ( \mP(1)[i, j] \right ) = 1
\end{displaymath}
this implies that
\begin{displaymath}
\sum_{i=1}^k \sum_{j=1}^k \left ( \mP(z)[i, j] \right ) = 1 \forall z \in [0, 1]
\end{displaymath}
\end{theorem}

\begin{proof}
Given a parameterised kernel $\mP(z) \in \mathbb{R}^{k \times k}$ we define
the sum function $\mathcal{S}_\mP(z)$ as
\begin{displaymath}
\mathcal{S}_\mP(z)) = \sum_{i=1}^k \sum_{j=1}^k \left ( \mP(z)[i, j] \right )
\end{displaymath}
As specified in Theorem \ref{th:sum_one} we assume that $\mathcal{S}_\mP(0)
= \mathcal{S}_\mP(1) = 1$ is given. From Lemma \ref{le:sum_of_affine_functions},
we know that $\mathcal{S}_\mP(z)$ is an affine function. We now show that
\begin{displaymath}
\mathcal{S}_\mP(0) = \mathcal{S}_\mP(1) = 1 \Rightarrow \mathcal{S}_\mP(z) = 1 \forall z \in [0, 1]
\end{displaymath}
by contradiction. Assume there is a $z' \in [0, 1]$ for which it holds that
$\mathcal{S}_\mP(z') = r$ where $r \in \mathbb{R}, r > 1$. There must then
be at least one subinterval $v_l \subseteq [0, z']$ on which
$\mathcal{S}_\mP(z)$ is strictly increasing, i.e. $\frac{d}{dz}
\mathcal{S}_\mP(z)) > 0$. Similarly, there must be at least one subinterval
$v_r \subseteq [z', 1]$ on which $\mathcal{S}_\mP(z)$ is strictly
decreasing, i.e. $\frac{d}{dz} \mathcal{S}_\mP(z) < 0$.\\

However, an affine
function $f(z) = mz + b$ has a constant derivative since $\frac{d}{dz} f(z)
= m$ which is independent of $z$. This implies that $\mathcal{S}_\mP(z)$ is
not an affine function which is a contradiction to our assumptions. There
can hence be no $z'$ for which $\mathcal{S}_\mP(z') > 1$. A similar
argument can be made for $r<1$, the proposition follows from these two
arguments.
\end{proof}

\section{Parameterised Kernels for Common Perturbations} \label{sec:appendix_parameterised_kernel_derivations}
Akin to the derivations in Section
\ref{ssec:parameterised_kernels_for_modelling_camera_shake} we here derive
parameterised kernels for box blur and sharpen perturbations as well as
motion blur perturbations with other blur angles. The derivations are done
for kernels of arbitrary uneven sizes, but to visualise the results in a
concise manner examples for small kernel sizes such as $s=3$ are shown.
\subsection{Generalised Camera Shake Kernels}
\label{ssec:generalised_camera_shake_kernels}
We also run tests for motion blur angles of $0, 90$ and $135$ degrees. The
definition of the center entry is the same for all of these angles. For
$135^\circ$ the values on the main diagonal become $\frac{1}{s}z$ with
the respective other entries still being $0$. For $0^\circ$ those in the
column crossing through the center and for $90^\circ$ those in the row
crossing through the center become $\frac{1}{s}z$ with the respective other
entries being zero.

The derivation shown in Example \ref{ex:motion_blur_3x3} can easily be
generalised to kernels of arbitrary sizes $s \in \mathbb{N}, s \geq 3$
where $s$ is an uneven number. In those cases we obtain that
\begin{align*}
p_\text{center}(z) &= \left ( \frac{1}{s} - 1 \right ) z + 1\\
p_\text{antidiagonal}(z) &= \frac{1}{s} z\\
p_\text{off-antidiagonal}(z) &= 0
\end{align*}

\subsection{Derivations for Parameterised Box Blur Kernels}
\label{ssec:derivations_for_parameterised_box_blur_kernels}
Box blur kernels are among the simplest kernels and are, for example,
introduced by \citet[pp.153--154]{ShapiroStockman01}. The two initial
conditions for a box blur kernel of size $s=3$ as presented in their work
are
\begin{displaymath}
\mP_{z=0} = \begin{pmatrix}
0 & 0 & 0 \\[2pt]
0 & 1 & 0 \\[2pt]
0 & 0 & 0 \\
\end{pmatrix},
\mP_{z=1} = \begin{pmatrix}
\frac{1}{9} & \frac{1}{9} & \frac{1}{9} \\[2pt]
\frac{1}{9} & \frac{1}{9} & \frac{1}{9} \\[2pt]
\frac{1}{9} & \frac{1}{9} & \frac{1}{9} \\
\end{pmatrix}
\end{displaymath}
\paragraph{Center Entry} For the center entry we have that
\begin{align}
p(0) = 1 = a \cdot 0 + b \label{eq:bblur1}\\
p(1) = \frac{1}{s^2} = a \cdot 1 + b \label{eq:bblur2}
\end{align}
From Equation \ref{eq:bblur1} we obtain $b=1$, using this in Equation
\ref{eq:bblur2} yields $\frac{1}{s^2} = a \cdot 1 + 1$ which implies $p(z)
= \left ( \frac{1}{s^2} - 1 \right ) z + 1$.
\paragraph{Non-center Entries} We know that
\begin{align}
p(0) = 0 = a \cdot 0 + b\\
p(1) = \frac{1}{s^2} = a \cdot 1 + b
\end{align}
so the first equation implies $b=0$. Using this, it follows from the second
equation that $a=\frac{1}{s^2}$. In summary, we obtain $p(z) =
\frac{1}{s^2} z$ for the non-center entries.
\paragraph{Box Blur Parameterisation} We have
\begin{align*}
p_\text{center}(z) &= \left ( \frac{1}{s^2} - 1 \right ) z + 1\\
p_\text{non-center}(z) &= \frac{1}{s^2} z\\
\end{align*}

\subsection{Derivations for Parameterised Sharpen Kernels}
\label{ssec:derivations_for_parameterised_sharpen_kernels}
For sharpen kernels we draw on the discussions by
\citet[pp.50--56]{Arvo91}. The initial conditions for a sharpen kernel are
\begin{displaymath}
\mP_{z=0} = \begin{pmatrix}
0 & 0 & 0 \\[2pt]
0 & 1 & 0 \\[2pt]
0 & 0 & 0 \\
\end{pmatrix},
\mP_{z=1} = \begin{pmatrix}
0 & -\frac{1}{4} & 0 \\[2pt]
-\frac{1}{4} & 2 & -\frac{1}{4} \\[2pt]
0 & -\frac{1}{4} & 0 \\
\end{pmatrix}
\end{displaymath}
We refer to the entry at the center of the kernel as the center entry.
Given a kernel size $s$ and a row index $j \in [1, \frac{s-1}{2}]$, we
refer to the first $\frac{s-1}{2} - j$ and the last $\frac{s-1}{2} - j$
entries in the row as zero entries. For row indices $j \in [\frac{s+1}{2},
s]$ we refer to the first $j - \frac{s+1}{2}$ and the last $j -
\frac{s+1}{2}$ entries in the row as zero entries. We further to refer to
all entries that are neither center nor zero entries as negative entries.
To make these definitions easier to understand we visualise the types of
entries for a $5 \times 5$ kernel with $e_c, e_z, e_n$ denoting center,
zero and negative entries, respectively:
\begin{displaymath}
\left (\begin{array}{ccccc}
\cellcolor{Orange!20} e_z & \cellcolor{Orange!20} e_z &
\cellcolor{OliveGreen!20} e_n & \cellcolor{Orange!20} e_z &
\cellcolor{Orange!20} e_z \\
\cellcolor{Orange!20} e_z & \cellcolor{OliveGreen!20} e_n &
\cellcolor{OliveGreen!20} e_n & \cellcolor{OliveGreen!20} e_n &
\cellcolor{Orange!20} e_z \\
\cellcolor{OliveGreen!20} e_n & \cellcolor{OliveGreen!20} e_n &
\cellcolor{NavyBlue!20} e_c & \cellcolor{OliveGreen!20} e_n &
\cellcolor{OliveGreen!20} e_n \\
\cellcolor{Orange!20} e_z & \cellcolor{OliveGreen!20} e_n &
\cellcolor{OliveGreen!20} e_n & \cellcolor{OliveGreen!20} e_n &
\cellcolor{Orange!20} e_z \\
\cellcolor{Orange!20} e_z & \cellcolor{Orange!20} e_z &
\cellcolor{OliveGreen!20} e_n & \cellcolor{Orange!20} e_z &
\cellcolor{Orange!20} e_z \\
\end{array} \right )
\end{displaymath}
For the sake of simplicity, we generalise the kernels presented in the
literature to larger kernel sizes by assuming that all negative entries in
the kernel have the same value. We can then compute the following affine
parameterisations for the kernel entries:

\paragraph{Center Entry} For the sharpen case it holds for the center entry
that
\begin{align}
p(0) = 1 = a \cdot 0 + b \label{eq:sharpen1}\\
p(1) = 2 = a \cdot 1 + b \label{eq:sharpen2}
\end{align}
$b=1$ immediately follows from the Equation \ref{eq:sharpen1} and Equation
\ref{eq:sharpen2} then implies that $a=1$, we arrive at $p(z) = z + 1$

\paragraph{Zero Entries} We know that $p(0) = p(1) = 0$ and therefore $p(z)
= 0$.

\paragraph{Negative Entries} Since each negative entry should have the same
value and the sum of all elements in the kernel should be one, we need to
know how many negative entries there are in a kernel of size $s \times s$.
To determine this, we first count the number of zero entries in one of the
four corners of the kernel. For the top-left corner in the $5 \times 5$
kernel for example, there are two zero entries in the first row and one in
the second row. In general, there are always $\frac{s-1}{2}$ zeros in the
first row, $\frac{s-1}{2} - 1$ zeros in the second row and so on until we
reach one zero entry in the $\frac{s-1}{2}$-th row. Each corner therefore
has $\frac{1}{2} \frac{s-1}{2} \left (\frac{s-1}{2}+1 \right )$ zero
entries and since there are four such corners there are $2
\frac{s-1}{2}\left ( \frac{s-1}{2}+1 \right )=:q_z$ zero entries in total.
The number of negative entries is hence the total number of entries minus
the one center entry and the number of zero entries, i.e. $s^2 - q_z - 1 =:
q_n$. We get that
\begin{align}
p(0) = 0 = a \cdot 0 + b \label{eq:sharpen3}\\
p(1) = -\frac{1}{q_n} = a \cdot 1 + b \label{eq:sharpen4}
\end{align}
Equation \ref{eq:sharpen3} implies $b=0$ and from Equation
\ref{eq:sharpen4} it follows that $b=-\frac{1}{q_n}$, hence $p(z) =
-\frac{1}{q_n} z$.

\paragraph{Sharpen Parameterisation} We obtain
\begin{align*}
p_\text{center}(z) &= z + 1\\
p_\text{zero}(z) &= 0\\
p_\text{negative}(z) &= -\frac{1}{q_n} z\\
\end{align*}

\section{Derivations for Even Kernel Sizes} \label{sec:appendix_even_kernel_sizes}
As mentioned before we omit experiments for even kernel sizes in this work
due to the fact that the identity kernel is not well-defined for such
kernel sizes. An example of an approximation for an evenly sized identity
kernel, here in the $s=4$ case, is given as
\begin{displaymath}
\begin{pmatrix}
0 & 0 & 0 & 0\\[2pt]
0 & \frac{1}{4} & \frac{1}{4} & 0 \\[2pt]
0 & \frac{1}{4} & \frac{1}{4} & 0 \\[2pt]
0 & 0 & 0 & 0\\[2pt]
\end{pmatrix}
\end{displaymath}
Convolving an input image with this kernel introduces a slight blurring
effect since it is effectively a $2 \times 2$ box blur kernel. For
high-resolution images the use of evenly-sized kernels might still be
possible since the effect is less visible there, but for low-resolution
images it is most likely infeasible. We show parameterisations for some
evenly-sized kernels of size $s \times s$ here, they can be derived using
the same methodology employed in Section
\ref{ssec:parameterised_kernels_for_modelling_camera_shake} and Appendix
\ref{sec:appendix_parameterised_kernel_derivations}. In those kernels we refer to the
entries at positions $\left (\frac{s}{2}, \frac{s}{2} \right ), \left
(\frac{s}{2}, \frac{s}{2} + 1 \right ), \left (\frac{s}{2} + 1, \frac{s}{2}
\right )$ and $\left (\frac{s}{2} + 1, \frac{s}{2} + 1 \right )$ as center
entries.

\subsection{Camera Shake}
\paragraph{Angle $\phi \in \left \{0^\circ, \phi \in 90^\circ \right \}$}
For a kernel size of $4$ and a motion blur angle of $0$ degrees the initial
conditions are
\begin{displaymath}
\mP_{z=0} = \begin{pmatrix}
0 & 0 & 0 & 0\\[2pt]
0 & \frac{1}{4} & \frac{1}{4} & 0 \\[2pt]
0 & \frac{1}{4} & \frac{1}{4} & 0 \\[2pt]
0 & 0 & 0 & 0\\[2pt]
\end{pmatrix},
\mP_{z=1} = \begin{pmatrix}
0 & \frac{1}{8} & \frac{1}{8} & 0 \\[2pt]
0 & \frac{1}{8} & \frac{1}{8} & 0 \\[2pt]
0 & \frac{1}{8} & \frac{1}{8} & 0 \\[2pt]
0 & \frac{1}{8} & \frac{1}{8} & 0 \\
\end{pmatrix}
\end{displaymath}
Since there is no true center row/column that we would need to model motion
blur, we use the two rows/columns that are closest to the true center. For
arbitrary even filter sizes we then derive that
\begin{align*}
p_\text{center}(z) &= \left ( \frac{1}{2s} - \frac{1}{4} \right ) z +
\frac{1}{4}\\
p_\text{non-center}(z) &= \frac{1}{2s} z\\
\end{align*}

\paragraph{Angle $\phi \in \left \{45^\circ, \phi \in 135^\circ \right \}$}
The initial conditions for a kernel size of $4$ and an angle of $\phi =
45^\circ$ are
\begin{displaymath}
\mP_{z=0} = \begin{pmatrix}
0 & 0 & 0 & 0\\[2pt]
0 & \cellcolor{Orange!20}\frac{1}{4} & \cellcolor{OliveGreen!20}
\frac{1}{4} & 0 \\[2pt]
0 & \cellcolor{OliveGreen!20} \frac{1}{4} & \cellcolor{Orange!20}
\frac{1}{4} & 0 \\[2pt]
0 & 0 & 0 & 0\\[2pt]
\end{pmatrix},
\mP_{z=1} = \left (\begin{array}{ccccc}
0 & 0 & 0 & \frac{1}{4} \\[2pt]
0 & 0 & \frac{1}{4} & 0 \\[2pt]
0 & \frac{1}{4} & 0 & 0 \\[2pt]
\frac{1}{4} & 0 & 0 & 0 \\
\end{array} \right )
\end{displaymath}
Unlike the $0^\circ$ and $90^\circ$ motion blur trails, the $45^\circ$ and
$135^\circ$ trails are well-defined in the case of even kernel sizes. This
means that the parameterisation is different for center entries that lie on
the antidiagonal in the $45^\circ$ case/main diagonal in the $135^\circ$
case and those center entries that do not. We refer to the center entries
on the respective diagonal as \textit{on-center}, marked in green in the
above example, and those center entries that do not lie on the diagonal as
\textit{off-center}, marked in orange in the above example. We obtain the
following parameterisation
\begin{align*}
p_\text{center}(z) &= \left ( \frac{1}{s} - \frac{1}{4} \right ) z +
\frac{1}{4}\\
p_\text{off-center}(z) &= -\frac{1}{4} z + \frac{1}{4}\\
p_\text{non-center}(z) &= \frac{1}{s} z
\end{align*}

\subsection{Box Blur}
The initial conditions for a $4 \times 4$ box blur kernel are
\begin{displaymath}
\mP_{z=0} = \begin{pmatrix}
0 & 0 & 0 & 0\\[2pt]
0 & \frac{1}{4} & \frac{1}{4} & 0 \\[2pt]
0 & \frac{1}{4} & \frac{1}{4} & 0 \\[2pt]
0 & 0 & 0 & 0\\[2pt]
\end{pmatrix},
\mP_{z=1} = \begin{pmatrix}
\frac{1}{16} & \frac{1}{16} & \frac{1}{16} & \frac{1}{16} \\[2pt]
\frac{1}{16} & \frac{1}{16} & \frac{1}{16} & \frac{1}{16} \\[2pt]
\frac{1}{16} & \frac{1}{16} & \frac{1}{16} & \frac{1}{16} \\[2pt]
\frac{1}{16} & \frac{1}{16} & \frac{1}{16} & \frac{1}{16} \\
\end{pmatrix}
\end{displaymath}
We therefore obtain
\begin{align*}
p_\text{center}(z) &= \left ( \frac{1}{s^2} - \frac{1}{4} \right ) z +
\frac{1}{4}\\
p_\text{non-center}(z) &= \frac{1}{s^2} z\\
\end{align*}
\subsection{Sharpen}
For sharpen kernels we have the following initial conditions for a $4
\times 4$ example:
\begin{displaymath}
\mP_{z=0} = \begin{pmatrix}
0 & 0 & 0 & 0\\[2pt]
0 & \frac{1}{4} & \frac{1}{4} & 0 \\[2pt]
0 & \frac{1}{4} & \frac{1}{4} & 0 \\[2pt]
0 & 0 & 0 & 0\\[2pt]
\end{pmatrix}
\end{displaymath}
\begin{displaymath}
\mP_{z=1} = \begin{pmatrix}
0 & -\frac{1}{8} & -\frac{1}{8} & 0 \\[2pt]
-\frac{1}{8} & \frac{1}{2} & \frac{1}{2} & -\frac{1}{8} \\[2pt]
-\frac{1}{8} & \frac{1}{2} & \frac{1}{2} & -\frac{1}{8} \\[2pt]
0 & -\frac{1}{8} & -\frac{1}{8} & 0 \\
\end{pmatrix}
\end{displaymath}
We once again have four center entries $e_c$. For a kernel size $s$ and an
index $j \in [1, \frac{s}{2}-1]$ we refer to the first $\frac{s}{2}-1 - j$
and last $\frac{s}{2}-1 - j$ entries in a row as zero entries $e_z$. For $j
\in \frac{s}{2}+1, s]$ the first $j - \frac{s}{2} - 1$ and the last $j -
\frac{s}{2} - 1$ entries are zero entries $e_z$. We declare all other
entries as negative entries $e_n$. In the case of a $6 \times 6$ kernel the
classification of the entries is as follows:
\begin{displaymath}
\left (\begin{array}{cccccc}
\cellcolor{Orange!20} e_z & \cellcolor{Orange!20} e_z &
\cellcolor{OliveGreen!20} e_n & \cellcolor{OliveGreen!20} e_n &
\cellcolor{Orange!20} e_z & \cellcolor{Orange!20} e_z \\
\cellcolor{Orange!20} e_z & \cellcolor{OliveGreen!20} e_n &
\cellcolor{OliveGreen!20} e_n & \cellcolor{OliveGreen!20} e_n &
\cellcolor{OliveGreen!20} e_n & \cellcolor{Orange!20} e_z \\
\cellcolor{OliveGreen!20} e_n & \cellcolor{OliveGreen!20} e_n &
\cellcolor{NavyBlue!20} e_c & \cellcolor{NavyBlue!20} e_c &
\cellcolor{OliveGreen!20} e_n & \cellcolor{OliveGreen!20} e_n \\
\cellcolor{OliveGreen!20} e_n & \cellcolor{OliveGreen!20} e_n &
\cellcolor{NavyBlue!20} e_c & \cellcolor{NavyBlue!20} e_c &
\cellcolor{OliveGreen!20} e_n & \cellcolor{OliveGreen!20} e_n \\
\cellcolor{Orange!20} e_z & \cellcolor{OliveGreen!20} e_n &
\cellcolor{OliveGreen!20} e_n & \cellcolor{OliveGreen!20} e_c &
\cellcolor{OliveGreen!20} e_n & \cellcolor{Orange!20} e_z \\
\cellcolor{Orange!20} e_z & \cellcolor{Orange!20} e_z &
\cellcolor{OliveGreen!20} e_n & \cellcolor{OliveGreen!20} e_n &
\cellcolor{Orange!20} e_z & \cellcolor{Orange!20} e_z \\
\end{array} \right )
\end{displaymath}
The number of zero fields in this case is $\frac{1}{2} \left ( \frac{s}{2}
- 1 \right ) \left ( \frac{s}{2} - 1 + 1 \right )$ per corner and therefore
their total number is $q_z = 2 \left ( \frac{s}{2} - 1 \right ) \left (
\frac{s}{2} \right )$. The number of negative fields is therefore $q_n =
s^2 - q_z - 4$, we derive that
\begin{align*}
p_\text{center}(z) &= \frac{1}{4} z + \frac{1}{4}\\
p_\text{negative}(z) &= -\frac{1}{q_n} z\\
p_\text{zero}(z) &= 0\\
\end{align*}

\end{document}